\documentclass[conference]{IEEEtran}
\IEEEoverridecommandlockouts
\usepackage{cite}
\usepackage{amsmath,amssymb,amsfonts}
\usepackage{graphicx}
\usepackage{textcomp}
\usepackage{booktabs}
\usepackage{url}
\usepackage{hyperref}
\def\BibTeX{{\rm B\kern-.05em{\sc i\kern-.025em b}\kern-.08em
    T\kern-.1667em\lower.7ex\hbox{E}\kern-.125emX}}

\begin{document}

\title{MoRAL: Sensor-Grounded BEV Reasoning for Compact VLMs\\
toward Edge-Oriented Autonomous Driving
}

\author{
\IEEEauthorblockN{Ambarish Govindarajulu Kaliamurthi}
\IEEEauthorblockA{
  \textit{Computer Engineering Department}\\
  \textit{San Jos\'{e} State University (SJSU)}\\
  San Jos\'{e}, CA, USA\\
  ambarish.govindarajulukaliamurthi@sjsu.edu}
\and
\IEEEauthorblockN{Kaikai Liu}
\IEEEauthorblockA{
  \textit{Computer Engineering Department}\\
  \textit{San Jos\'{e} State University (SJSU)}\\
  San Jos\'{e}, CA, USA\\
  kaikai.liu@sjsu.edu}
}

\maketitle
\begin{abstract}
Deploying vision-language models (VLMs) for safety-critical spatial
reasoning on resource-constrained autonomous driving platforms requires
both compact model size and reliable metric grounding.
We present MoRAL (Multimodal Reasoning for Autonomous Language Models),
a two-stage fine-tuning pipeline that teaches Cosmos-Reason2-2B~\cite{b14}
to first \emph{read} a physics-encoded Bird's Eye View (BEV) representation
and then \emph{reason} over it for driving decisions.
The BEV image encodes LiDAR metric distance as color bands, object class
as cluster morphology, and radar Doppler velocity as directional wedge
overlays, externalizing spatial perception into the input image so that
no learned 3D backbone is required at inference.
Stage~1 fine-tunes the vision encoder on 60{,}000 grounding records;
zero-shot baselines produce no parseable BEV outputs, confirming the
vocabulary requires explicit training.
Stage~2 fine-tunes the full model (52M parameters, 2.4\% of total) on
57{,}696 chain-of-thought records generated by Cosmos-Reason2-8B as
teacher, spanning eight driving question types.
On 2{,}304 held-out nuScenes frames evaluated by Gemma~4 (31B)~\cite{b16}
calibrated against human review, MoRAL wins seven of eight question types
over a zero-shot 8B baseline despite using four times fewer parameters,
with the largest margins on question types requiring structured multi-step
physics reasoning.
Emergency braking recall improves from 10.8\% to 47.8\%, output
degeneration falls from 94.1\% to 20.8\%, and the full pipeline fits a
consumer 8\,GB GPU at 42\,tok/s without quantization.
These results establish a reproducible foundation for compact,
physics-grounded VLM reasoning on mobile edge platforms.
\end{abstract}


\section{Introduction}

On-vehicle computing platforms impose strict constraints on model size,
inference latency, and pipeline complexity~\cite{b2}.
Safety-critical spatial reasoning---reading distances, velocities, and
collision trajectories from sensor data---must operate within these
constraints while remaining reliable.
Current multi-sensor perception systems address this through learned BEV
construction stages: multi-view camera backbones, spatial cross-attention,
and learned sensor fusion before any downstream reasoning~\cite{b17}.
These pipelines are effective but introduce inference-time complexity
that is difficult to accommodate on resource-constrained mobile platforms.

Large VLMs offer an alternative: given sufficient spatial grounding, a
single compact model could replace multiple perception stages.
The obstacle is that VLMs consistently fail at metric spatial reasoning.
DriveBench~\cite{b3} demonstrated this across 19{,}200 frames and 12
popular VLMs---models generate plausible driving responses even when all
visual input is removed, revealing reliance on language priors rather than
sensor geometry.
A systematic survey confirms that metric spatial grounding remains the
persistent bottleneck~\cite{b2}.

Scaling model size does not resolve this.
In our evaluation, a zero-shot 8B model recalls only 10.8\% of emergency
braking scenarios and defaults to empty or templated responses when
presented with an engineered BEV input.
The failure is representational, not parametric: the model cannot read
the spatial vocabulary encoded in sensor imagery regardless of its
language generation capacity.

The fix we explore is to move the perception burden into the input image
rather than the model.
If a BEV image explicitly encodes metric distance, object class, and
Doppler velocity as color and shape features, a compact model can be
trained to read and reason over them without a learned perception backbone
at inference.
The model's job shifts from perceiving the scene to interpreting a
pre-encoded spatial vocabulary---which turns out to be a learnable task
even at 2B parameters, as the results show.

This paper presents \textbf{MoRAL} (Multimodal Reasoning for Autonomous
Language Models), a two-stage fine-tuning pipeline for
Cosmos-Reason2-2B~\cite{b14} on nuScenes~\cite{b15}.
The contributions are:
\begin{enumerate}
\item A deterministic BEV rendering pipeline fusing LiDAR and radar into
896$\times$896\,px top-down images encoding distance, class, and velocity
as readable visual features.
\item A two-stage LoRA procedure: Stage~1 trains the vision encoder
(14.4M parameters) to decode the BEV vocabulary; Stage~2 trains the full
model (52M parameters, 2.4\% of 2.2B) on 57{,}696 teacher-generated
chain-of-thought records.
\item A four-condition evaluation with human-piloted judge calibration
showing MoRAL wins seven of eight question types over a zero-shot 8B
baseline, emergency braking recall improving from 10.8\% to 47.8\%.
\item Edge deployment validation confirming the full pipeline runs on a
consumer 8\,GB GPU at 4.3\,GB VRAM and 42\,tok/s without quantization.
\end{enumerate}

Extended results including a 42-condition ablation, per-question
consistency analysis, and think-block diagnostics are in the companion
thesis~\cite{b20}.

\section{Related Work}
\label{sec:related}

\subsection{VLMs for Autonomous Driving}

Language model adoption in autonomous driving has progressed through four
stages~\cite{b19}: scene explainers, modular VLA systems, end-to-end VLA
pipelines, and chain-of-thought augmented systems.
DriveGPT4~\cite{b4} pioneered decision explanation from front-camera video.
DriveVLM~\cite{b5} identified metric spatial grounding as a persistent
failure mode.
DriveLM~\cite{b6} established multi-question evaluation on nuScenes, which
MoRAL extends to eight physics-grounded question types.
OmniDrive~\cite{b18} highlighted counterfactual reasoning, motivating
our Q8 type.
Recent systems such as OpenDriveVLA~\cite{b7} and DriveWorld-VLA~\cite{b8}
pursue closed-loop trajectory planning through full 3D world-model
integration; MoRAL does not compete on closed-loop benchmarks.

\subsection{BEV-Based Language Reasoning}

Talk2BEV~\cite{b9} demonstrated language-BEV interaction for scene
understanding; BEVDriver~\cite{b10} showed BEV-LLM integration for
closed-loop driving.
VLA-MP~\cite{b11} is the closest published system: it integrates
multimodal BEV perception via ResNet50 and PointPillar backbones with
BEVFusion~\cite{b17}, a Q-Former projector, LLaVA-7B, and a GRU-bicycle
dynamics adapter for trajectory output.
MoRAL differs along four axes: (1)~\textit{goal}---safety-critical
reasoning rather than trajectory execution; (2)~\textit{BEV
construction}---deterministic rendering rather than a learned 3D backbone;
(3)~\textit{physics}---explicit TTC and braking distance in training
targets; (4)~\textit{scale}---a fine-tuned 2B model outperforming a
zero-shot 8B baseline on the spatial reasoning tasks evaluated here.
Direct metric comparison with VLA-MP is not meaningful: it targets
closed-loop trajectory execution on LangAuto/CARLA, while MoRAL targets
open-loop safety reasoning on nuScenes.

\subsection{Evaluation of Reasoning and Parameter-Efficient Fine-Tuning}

Evaluating reasoning quality in driving VLMs is an open problem.
DriveBench~\cite{b3} showed standard metrics can be fooled by fluent but
sensor-ungrounded responses.
Reason2Drive~\cite{b12} demonstrated that chain-of-thought quality is
measurable through domain-specific training with structured intermediate
supervision.
The nuScenes benchmark~\cite{b15} supports large-scale scene understanding
evaluation, but existing VQA protocols on it do not assess multi-step
physics reasoning or safety-critical decision quality.
LLM-as-judge evaluation has been adopted in recent reasoning
benchmarks~\cite{b2,b12} as a scalable alternative to exhaustive human
annotation, requiring the judge to be calibrated against human review and
independent of the model under evaluation; MoRAL follows both requirements.
LoRA~\cite{b13} reduces trainable parameters by injecting low-rank matrices
into frozen layers: given $W_0 \in \mathbb{R}^{d \times k}$, the update is
$W = W_0 + BA$ where $r \ll \min(d,k)$, reducing parameters from $dk$ to
$r(d+k)$ per layer.

\section{MoRAL System Design}
\label{sec:system}

\subsection{System Overview}

\begin{figure}[t]
  \centering
  \includegraphics[width=\columnwidth]{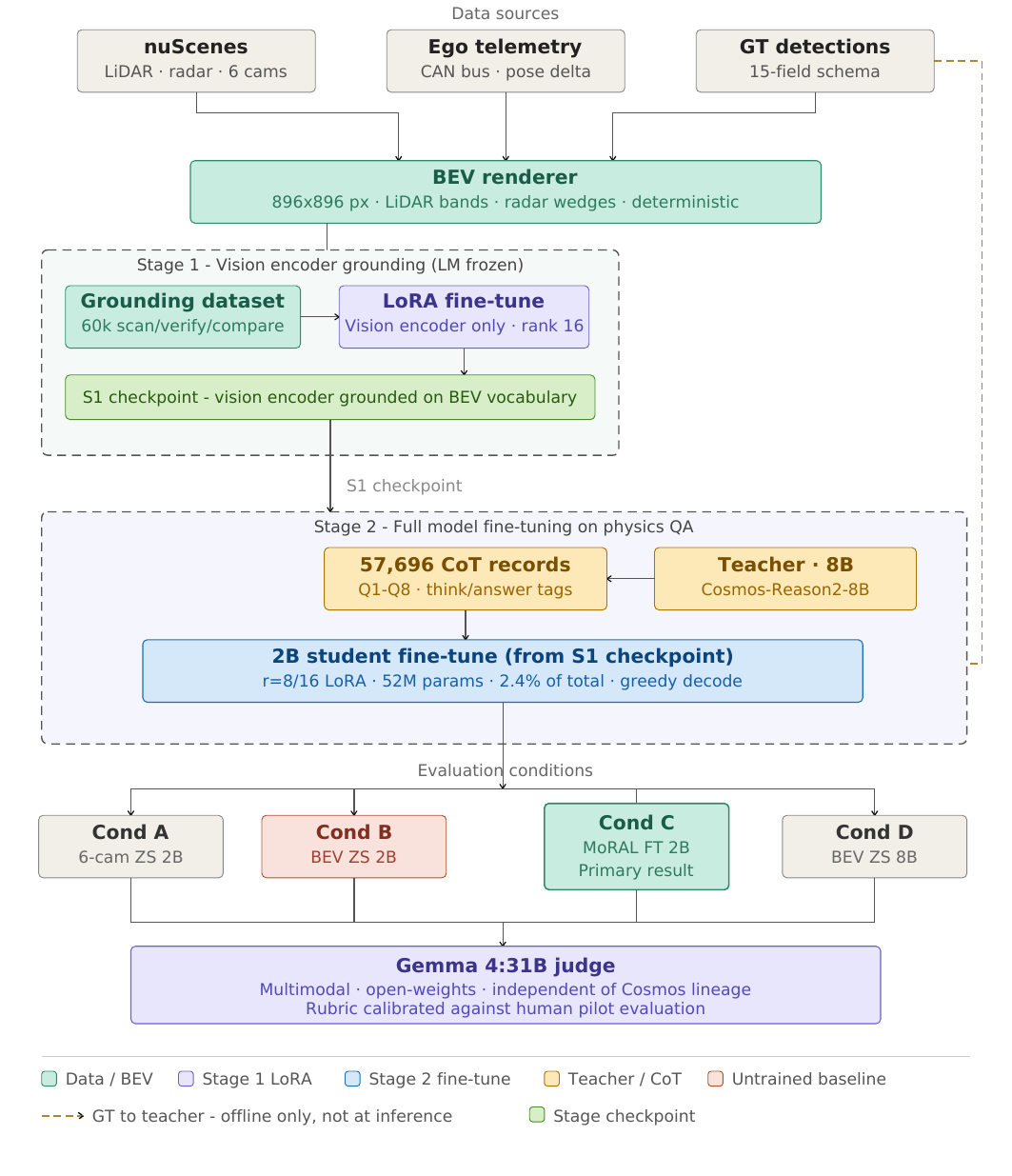}
  \caption{MoRAL two-stage pipeline. A deterministic BEV renderer fuses
           nuScenes LiDAR and radar into a physics-encoded top-down image.
           Stage~1 teaches BEV vocabulary decoding (LM frozen). Stage~2
           trains reasoning over eight driving question types using an 8B
           teacher. Four conditions isolate effects of modality and
           fine-tuning.}
  \label{fig:pipeline}
\end{figure}

MoRAL is a two-stage supervised fine-tuning pipeline applied to
Cosmos-Reason2-2B~\cite{b14}, built on the Qwen3-VL architecture with
native \texttt{<think>} and \texttt{<answer>} token support.
Every conclusion must trace back to a visible BEV feature; physics values
are precomputed for the 8B teacher but withheld from the 2B student at
inference.
Fig.~\ref{fig:pipeline} illustrates the pipeline.

The two-stage structure arose from empirical failure: training Stage~2
reasoning on the base model produced a model that copied numeric values
from text supplements rather than reading the BEV image; Stage~1 was
introduced to establish the visual vocabulary first.
Cosmos-Reason2 was selected over Qwen2.5-VL and Qwen3-VL primarily
because its post-training targets physical-world reasoning rather than
general instruction following.
Preliminary comparisons on spatial physics questions showed more consistent
collision trajectory reasoning than general-purpose alternatives.
The shared backbone between the 8B teacher and 2B student also enables
direct \texttt{<think>...<answer>} format transfer without prompt
reformatting.

\subsection{BEV Sensor Representation}

\begin{figure}[t]
  \centering
  \includegraphics[width=\columnwidth]{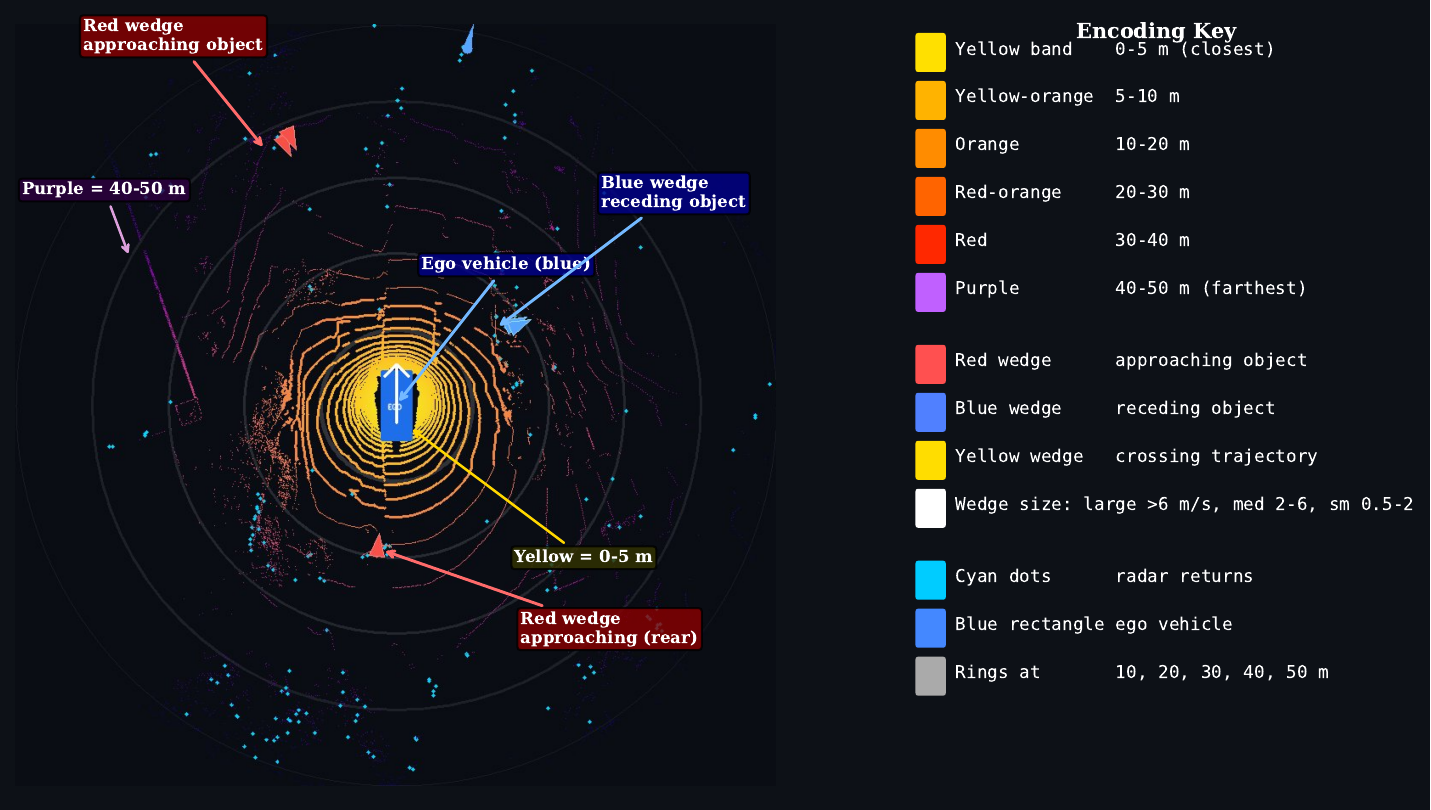}
  \caption{BEV sensor representation. LiDAR color bands encode metric
           distance (yellow=near, purple=far). Radar wedges encode velocity
           direction and closing speed. Range rings at 10--50\,m provide
           metric reference.}
  \label{fig:bev}
\end{figure}

The BEV image is a 896$\times$896\,px top-down rendering of a
100$\times$100\,m ego-centric region (Fig.~\ref{fig:bev}).
The design was informed by earlier ablations: arrow-style radar overlays
caused post-fine-tuning degradation due to geometric conflict with learned
cluster shapes, and bounding-box annotations led to hallucination at
clean-BEV inference; the final design uses filled wedge overlays and clean
LiDAR point clouds.
\textbf{LiDAR distance bands:} returns are colored by metric distance
(yellow 0--5\,m through purple 40--50\,m); cluster morphology encodes
class (wide-dense for vehicles, elongated for barriers, compact-sparse for
pedestrians). Ring accuracy degrades beyond 30\,m due to sparsity of the
32-beam Velodyne HDL32E at far range.
\textbf{Radar Doppler wedges:} each Continental ARS408 detection with a
reliable quality flag is rendered as a filled triangle---color encodes
direction (red approaching, blue receding, yellow crossing), size encodes
closing speed (large $>$6\,m/s, medium 2--6\,m/s, small 0.5--2\,m/s).
\textbf{Ego telemetry:} speed, heading, yaw rate, and steering angle from
the CAN bus are provided as structured text.
The renderer is fully deterministic at inference; the vertical axis is
discarded, so objects at different elevations may produce ambiguous
clusters.

\subsection{Evaluation Conditions}

Four conditions form a controlled comparison (Table~\ref{tab:conditions}).
Stage~1 confirmed neither base model reliably decodes the BEV vocabulary
zero-shot: the 2B produces 0/808 parseable responses; the 8B defaults to
empty on 79.2\% of frames.
MoRAL (Condition~C) is therefore the only condition whose BEV reading
capability was empirically validated before Stage~2 evaluation; Conditions
B and D are best understood as language-model priors conditioned on a
partially unread sensor image.
Condition~A used a camera-only prompt since it receives no BEV input.

\begin{table}[t]
\caption{Evaluation Conditions}
\label{tab:conditions}
\begin{center}
\begin{tabular}{llll}
\toprule
\textbf{Cond.} & \textbf{Model} & \textbf{Input} & \textbf{Role} \\
\midrule
A & 2B zero-shot  & 6 cameras    & Camera baseline \\
B & 2B zero-shot  & BEV + CAM\_F & Untrained BEV \\
C & 2B fine-tuned & BEV + CAM\_F & MoRAL (ours) \\
D & 8B zero-shot  & BEV + CAM\_F & Scale baseline \\
\bottomrule
\end{tabular}
\end{center}
\end{table}

\section{Dataset Construction and Training}
\label{sec:training}

\subsection{Stage~1: BEV Grounding}

A quality scoring function evaluated approximately 34{,}000 nuScenes
training frames on six criteria: LiDAR color-band diversity
(weight 3.0$\times$), class diversity (1.5$\times$), detection count
4--25 (1.0$\times$), ego speed $>$0.5\,m/s (1.0$\times$), radar wedge
presence (1.0$\times$), and visibility quality (0.5$\times$).
The top 10{,}000 frames were retained.
Three record types were generated: scan (complete eight-zone BEV
description), verify (single-zone confirmation, anti-hallucination), and
compare (two-zone distance ranking).
The dataset contains 60{,}000 records split 49{,}400/4{,}945/5{,}655
train/val/test.
LoRA rank~16 is applied to vision encoder attention layers only
(14.4M trainable parameters, LM frozen), learning rate 2e-4, cosine decay;
the epoch-3 checkpoint was selected based on validation Zone~F1.

\subsection{Stage~2: Physics QA}

\textbf{Anti-shortcut schema.}
Raw nuScenes detections (30+ fields) are reduced to a 15-field schema:
distance to six bands, heading to four categories, velocity retained only
when radar quality is ``reliable,'' preventing ground-truth leakage through
the detection supplement.

\textbf{Physics precomputation.}
Braking distances ($d_\mathrm{dry} = v^2_\mathrm{ego}/8.0$,
$d_\mathrm{wet} = v^2_\mathrm{ego}/4.0$) and TTC
($= d_\mathrm{obj}/v_\mathrm{closing}$) are computed for the teacher and
withheld from the student at inference.
Required action: EMERGENCY\_BRAKE if TTC~$<$~1.5\,s, BRAKE if~$<$~3\,s,
MONITOR if~$<$~5\,s, MAINTAIN otherwise.

\textbf{Question types and training.}
Eight question types (Q1--Q8): approach rate, zone grounding, threat
ranking, blind spots, driving decision, sensor uncertainty, safety ethics,
counterfactual reasoning.
Records enforce ordered tags \texttt{[CONDITIONS]}, \texttt{[OBSERVE\_BEV]},
\texttt{[PHYSICS\_WORK]}, \texttt{[SECOND\_CHECK]}, \texttt{[UNCERTAINTY]};
a response placing \texttt{[PHYSICS\_WORK]} before \texttt{[OBSERVE\_BEV]}
is rejected, ensuring observations precede physics computation.
Generated by Cosmos-Reason2-8B via vLLM (temperature 0.7); 57{,}696 of
$\sim$70{,}000 passed 11 hard rejection criteria including empty response,
misordered tags, ground-truth leaks, and per-question structural checks.
Fine-tuned from the Stage~1 checkpoint with LoRA rank~8 on the vision
encoder and rank~16 on all LM linear layers ($\sim$52M parameters, 2.4\%
of 2.2B), learning rate 1e-4, cosine decay, max sequence length 4{,}096,
bfloat16, single H100 PCIe 80\,GB.
Greedy decoding with repetition penalty 1.3 is used at inference.

\section{Validation Results}
\label{sec:eval}

\subsection{Evaluation Protocol}

All evaluation uses the nuScenes validation split: 2{,}304 records across
eight question types on 2{,}042 unique frames, with training data drawn
exclusively from the training split.
The ground-truth action distribution is strongly imbalanced (85.5\%
MAINTAIN, 1.6\% EMERGENCY\_BRAKE across 68{,}082 objects), making standard
detection metrics such as mAP inapplicable: mAP requires localization
predictions matched by IoU, which MoRAL does not produce, and aggregate
accuracy would be dominated by MAINTAIN, rendering safety-critical cases
invisible.
Evaluation is structured around metrics that directly operationalize the
three claims: Zone~F1 and within\_20pct for vocabulary learnability,
per-question judge scores for reasoning quality, and EMERGENCY\_BRAKE
recall plus degeneration rate for safety-relevant behavior.

\textbf{Human evaluation and transition to LLM-as-judge.}
Initial evaluation was conducted by manual human review.
Annotating a single frame across eight question types required
cross-referencing the BEV map, six camera views, and the detections
schema, taking 10--20 minutes per frame.
A 40-frame pilot study (10 per condition) required several hours of focused
effort and established a baseline condition ranking and failure mode
characterization.
Extrapolating to the full 2{,}304-record set would require hundreds of
hours of expert annotation, making exhaustive manual review impractical.
We therefore adopted Gemma~4 (31B)~\cite{b16} as the primary evaluator,
selected for multimodal input support, open weights, local reproducibility,
and independence from the Cosmos model family.
Gemma~4's separated thinking mode was key to calibration: the judge's
reasoning trace was inspectable, allowing iterative rubric refinement over
40--80 frames until judge rankings matched human review.
The final prompt includes eight guardrails penalizing unsupported spatial
claims, incorrect action labels, and physically inconsistent reasoning.
Because Gemma penalizes formatting failures more consistently than a human
reviewer under time pressure, reported scores are conservative relative to
human judgment.
Scores are normalized to $[0,1]$ and the composite is the unweighted mean
across all $N$ records:
\begin{equation}
  \mathrm{composite} = \frac{1}{N}\sum_{i=1}^{N} \frac{s_{\mathrm{raw},i}}{5}
  \label{eq:composite}
\end{equation}

\subsection{Stage~1: BEV Vocabulary Learnability}

Stage~1 is evaluated on 808 held-out validation frames.
\textbf{Zone~F1} measures binary zone occupancy prediction quality across
eight BEV zones (precision = correctly predicted occupied / all predicted
occupied; recall = correctly predicted occupied / all truly occupied).
\textbf{Within\_20pct} measures the fraction of occupied-zone distance
predictions falling within 20\% of ground-truth distance, testing whether
the color-band vocabulary translates to usable metric estimates.

\begin{table}[t]
\caption{Stage~1 BEV Vocabulary Readability (808 Validation Frames)}
\label{tab:stage1}
\begin{center}
\begin{tabular}{lccccc}
\toprule
\textbf{Model} & \textbf{Parse} & \textbf{Empty} & \textbf{Unique outputs} & \textbf{Zone F1} & \textbf{w20\%} \\
\midrule
Base 2B (ZS) & 0/808   & {---}  & 53/694  & N/A  & N/A  \\
Base 8B (ZS) & 645/808 & 79.2\% & 15/808  & N/A  & N/A  \\
MoRAL S1     & 803/808 & 0.4\%  & 781/808 & \textbf{0.89} & \textbf{0.58} \\
\bottomrule
\multicolumn{6}{l}{\footnotesize ZS=zero-shot, w20\%=within\_20pct gate metric}
\end{tabular}
\end{center}
\end{table}

MoRAL Stage~1 achieves Zone~F1 of 0.89 (precision 0.91, recall 0.90),
class accuracy 0.62, and velocity direction accuracy 0.89.
Near-range ring accuracy is strong: 0.788 (0--5\,m), 0.697 (5--10\,m),
0.710 (10--20\,m); accuracy collapses beyond 30\,m (0.227 at 30--40\,m,
0.039 at 40--50\,m), consistent with point cloud sparsity at far range
for the 32-beam Velodyne.
The within\_20pct gate reaches 0.58, marginally below 0.60; training
proceeded to Stage~2 because near-range accuracy covers the distance range
critical for emergency braking decisions.
Both zero-shot baselines fail entirely: the 2B produces 0 parseable
outputs with only 53 unique strings; the 8B defaults to empty on 79.2\%
of frames with only 15 distinct outputs.
The BEV vocabulary cannot be read without explicit training, validating
the two-stage design and ensuring Stage~2 gains in Condition~C are
attributable to a model that has actually learned to decode the sensor
interface.

\subsection{Grounded Reasoning Improves Performance (Claim~2)}

\begin{table}[t]
\caption{Per-Question Normalized Scores ($s_\text{raw}/5$) and Composite.
         $^*$~=~best per question.}
\label{tab:perq}
\begin{center}
\begin{tabular}{lcccc}
\toprule
\textbf{Question} & \textbf{A} & \textbf{B} & \textbf{C} & \textbf{D} \\
                  & \footnotesize{6-cam} & \footnotesize{BEV} & \footnotesize{BEV FT} & \footnotesize{BEV} \\
                  & \footnotesize{ZS 2B} & \footnotesize{ZS 2B} & \footnotesize{2B} & \footnotesize{ZS 8B} \\
\midrule
Q1 Approach Rate  & 0.361 & 0.241 & 0.533$^*$ & 0.362 \\
Q2 Grounding      & 0.323 & 0.227 & 0.432$^*$ & 0.353 \\
Q3 Threat Ranking & 0.352 & 0.243 & 0.502$^*$ & 0.373 \\
Q4 Blind Spots    & 0.374 & 0.287 & 0.534      & 0.552$^*$ \\
Q5 Decision       & 0.450 & 0.287 & 0.555$^*$ & 0.399 \\
Q6 Uncertainty    & 0.247 & 0.256 & 0.722$^*$ & 0.588 \\
Q7 Safety Ethics  & 0.347 & 0.304 & 0.509$^*$ & 0.481 \\
Q8 Counterfactual & 0.396 & 0.268 & 0.705$^*$ & 0.407 \\
\midrule
\textbf{Composite} & 0.362 & 0.266 & \textbf{0.565} & 0.439 \\
Judge avg (1--5)   & 1.81  & 1.33  & \textbf{2.83}  & 2.19  \\
\bottomrule
\end{tabular}
\end{center}
\end{table}

\begin{figure}[t]
  \centering
  \includegraphics[width=\columnwidth]{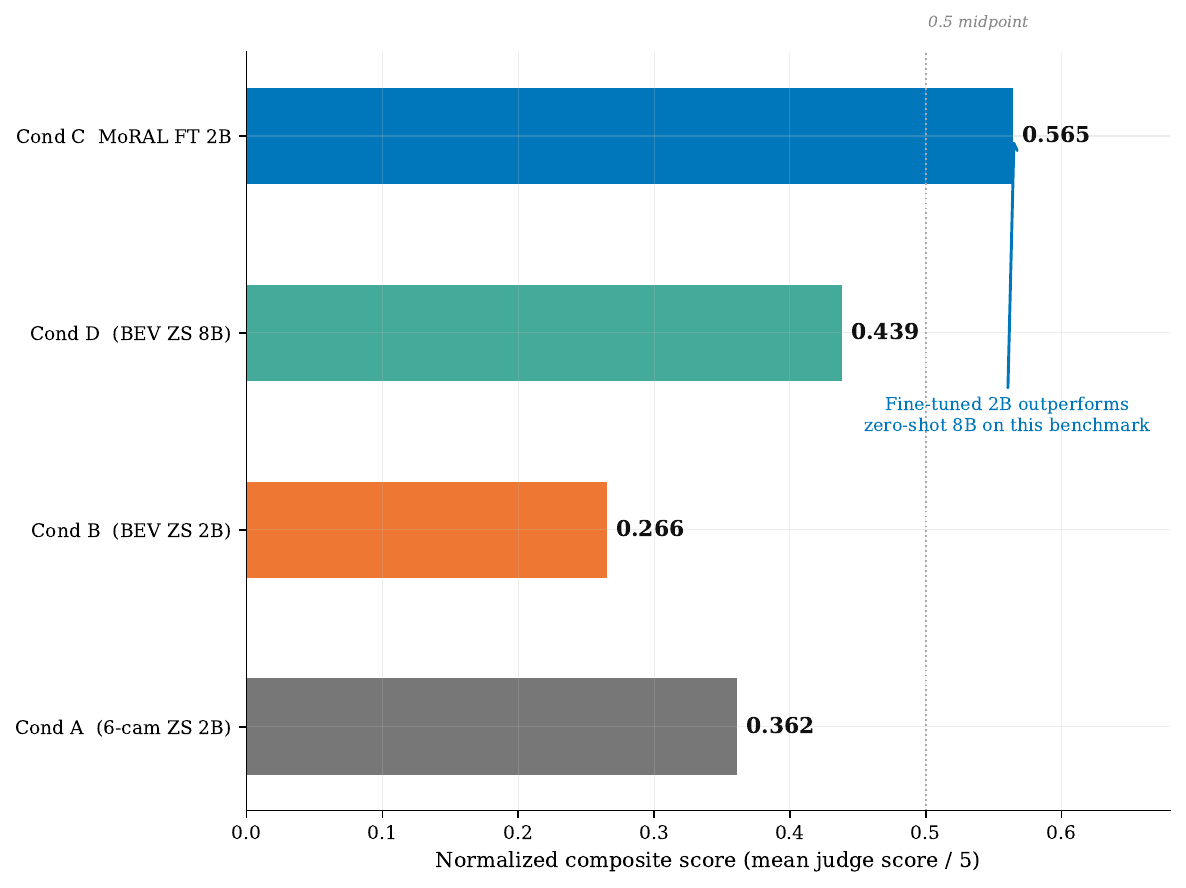}
  \caption{Composite scores across four conditions. MoRAL (Cond.~C, 0.565)
           outperforms the zero-shot 8B (Cond.~D, 0.439). Unlearned BEV
           (Cond.~B, 0.266) degrades below cameras only (Cond.~A, 0.362).}
  \label{fig:composite}
\end{figure}

\begin{figure}[t]
  \centering
  \includegraphics[width=\columnwidth]{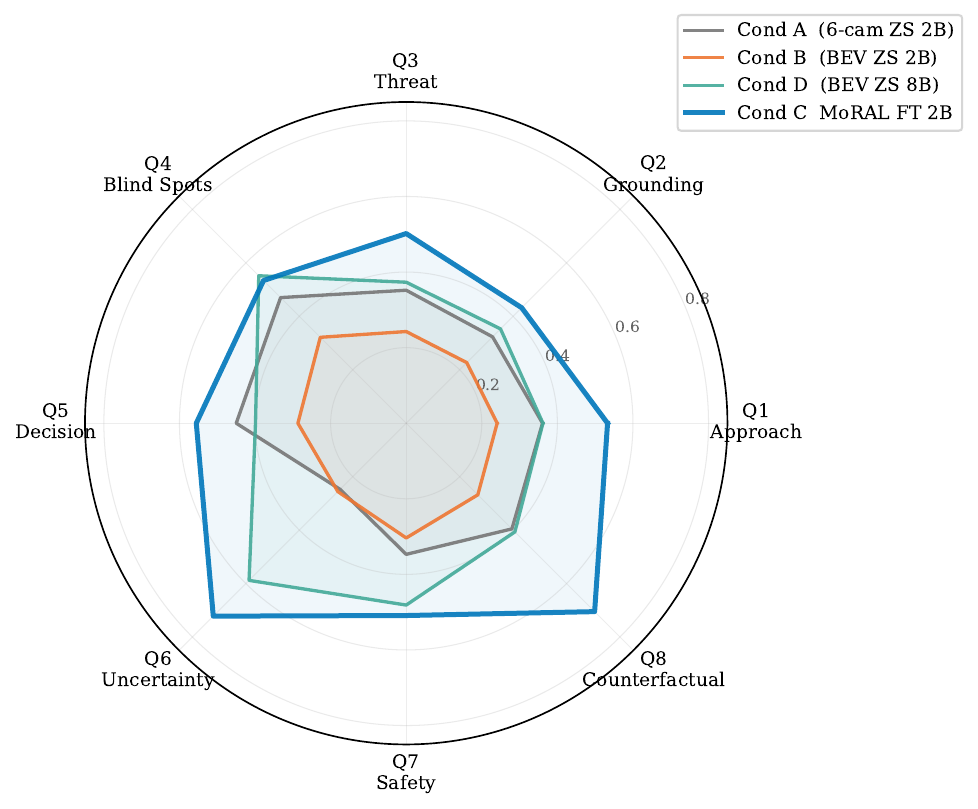}
  \caption{Per-question score profile. MoRAL (Cond.~C) dominates on Q1,
           Q3, Q5, Q6, Q8; Cond.~D is competitive only on Q4. Cond.~B
           traces the smallest area, reflecting near-total degeneration.}
  \label{fig:radar}
\end{figure}

MoRAL (Condition~C) achieves a composite score of 0.565
(Fig.~\ref{fig:composite}), a 29\% gain over the zero-shot 8B baseline
(0.439), leading on seven of eight question types (Fig.~\ref{fig:radar}).
The sole exception is Q4 (blind spots), where the 8B model's stronger
free-text generation for qualitative uncertainty provides an advantage.
The largest margins are on Q6 uncertainty (+0.134), Q8 counterfactual
(+0.298), and Q5 decision (+0.156)---the question types requiring
structured multi-step physics reasoning over specific sensor cues, where
BEV-grounded chain-of-thought training provides the most benefit.
Providing an unlearned BEV image to an untrained model (Condition~B)
reduces performance relative to cameras alone (0.266 vs.\ 0.362),
confirming MoRAL's gains come from the learned vocabulary, not BEV
input alone.
The 40-frame human pilot (Table~\ref{tab:human}) confirmed the same
ranking: MoRAL 3.18, Condition~A 1.70, Condition~D 1.66, Condition~B
1.13; Gemma's stricter rubric suggests reported scores are conservative.

\begin{table}[t]
\caption{Human Pilot Evaluation (10 Frames/Condition, 1--5 Scale)}
\label{tab:human}
\begin{center}
\begin{tabular}{lcp{3.8cm}}
\toprule
\textbf{Condition} & \textbf{Score} & \textbf{Primary Failure Mode} \\
\midrule
B: BEV ZS 2B   & 1.13 & Non-English output, format collapse \\
D: BEV ZS 8B   & 1.66 & Hallucinated spatial claims \\
A: 6-cam ZS 2B & 1.70 & Weak metric and physics reasoning \\
C: MoRAL FT 2B & \textbf{3.18} & Over-cautious on safety/ethics \\
\bottomrule
\end{tabular}
\end{center}
\end{table}

\subsection{Grounding Improves Safety Behavior (Claim~3)}

\begin{table}[t]
\caption{Action Safety Bias, Emergency Brake Recall, and Degeneration}
\label{tab:safety}
\begin{center}
\begin{tabular}{lcccc}
\toprule
\textbf{Condition} & \textbf{Correct} & \textbf{Sev.~Err.} & \textbf{EB Recall} & \textbf{Degen.} \\
\midrule
B: ZS 2B  & 24.8\% & $-$0.445 & 26.6\% & 94.1\% \\
C: MoRAL  & 30.6\% & $+$0.347 & \textbf{47.8\%} & \textbf{20.8\%} \\
D: ZS 8B  & 22.3\% & $-$0.788 & 10.8\% & 52.2\% \\
\bottomrule
\end{tabular}
\end{center}
\end{table}

\begin{figure}[t]
  \centering
  \includegraphics[width=\columnwidth]{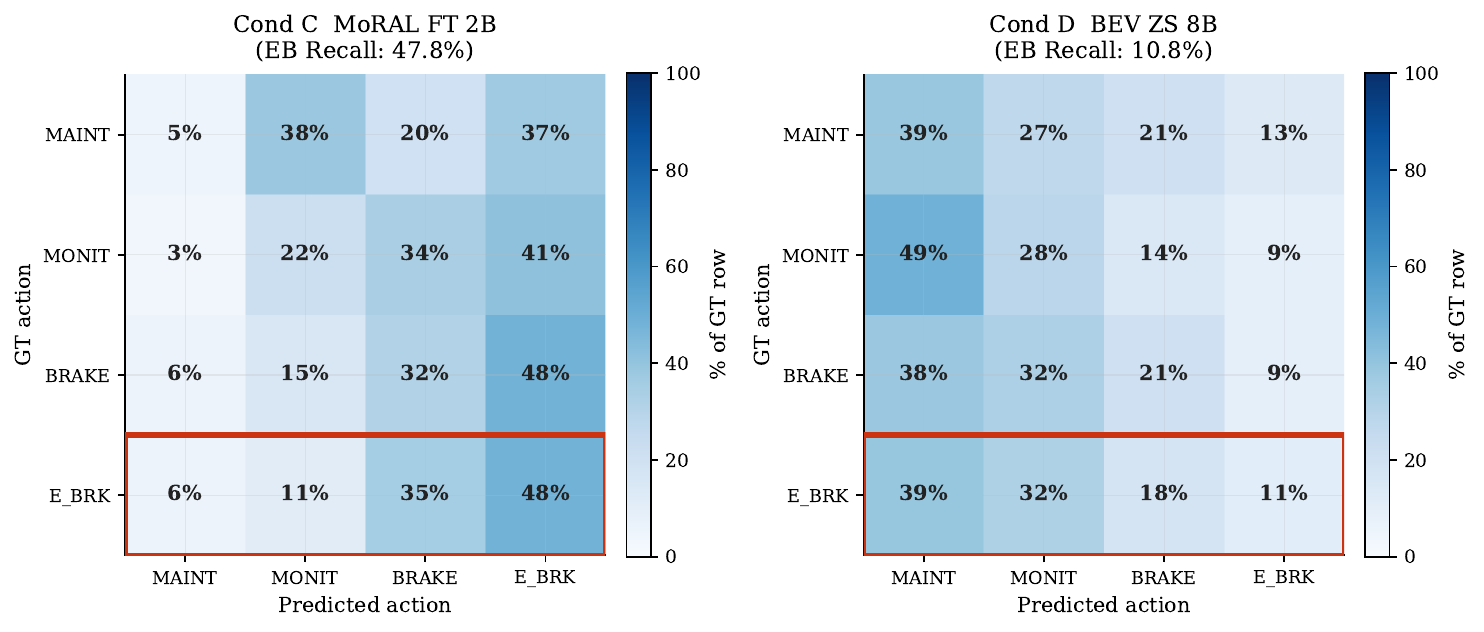}
  \caption{EMERGENCY\_BRAKE confusion matrices. Cond.~D under-reacts on
           60\% of Q5 records (severity $-$0.79); Cond.~C shifts toward
           over-reaction with 47.8\% EB recall.}
  \label{fig:confusion}
\end{figure}

The zero-shot 8B (Condition~D) systematically under-reacts: 60.0\% of Q5
records predict less severe actions than required, mean severity error
$-$0.79, recalling only 10.8\% of emergency braking cases.
Severity error is defined on a 0--3 scale (MAINTAIN=0 through
EMERGENCY\_BRAKE=3); negative values indicate under-reaction, positive
values indicate over-reaction.
In open-loop evaluation, systematic under-reaction is the more dangerous
pattern: a model that consistently misses imminent collisions provides no
safety benefit regardless of output fluency.
MoRAL shifts toward over-reaction (severity $+$0.35) with 47.8\% EB
recall (77 of 161 ground-truth EMERGENCY\_BRAKE records in Q5),
illustrated in Fig.~\ref{fig:confusion}.
Both patterns are relative comparisons on an open-loop benchmark; neither
represents acceptable production performance, and 47.8\% recall still
leaves most critical cases undetected.
Output degeneration falls from 94.1\% (Condition~B) to 20.8\%; the 8B
baseline at 52.2\% confirms scale alone does not resolve BEV reading
failures.

\subsection{Edge Deployment Validation}

The complete pipeline was executed locally on a consumer NVIDIA RTX 4070
Laptop GPU (8.2\,GB VRAM) without cloud infrastructure.
Table~\ref{tab:edge} reports measured metrics across all eight question
types on nuScenes validation frames.
The model peaks at 4.61\,GB VRAM and sustains 42\,tok/s in bfloat16
without quantization, producing 100\% English structured output across all
question types; the base model without fine-tuning produces non-English or
unparseable outputs on the same BEV inputs.
MoRAL replaces learned BEV construction stages with a deterministic
rendering step and a single 2B forward pass; the claim is not that it is
definitively faster end-to-end, but that it avoids a category of
inference-time component that systems such as BEVFusion~\cite{b17} require.
The 5.7\,s per-frame figure covers model inference only; BEV rendering is
deterministic CPU-side work not profiled here and is left for end-to-end
latency benchmarking in future work.

\begin{table}[t]
\caption{Edge Deployment Metrics --- NVIDIA RTX 4070 Laptop GPU}
\label{tab:edge}
\begin{center}
\begin{tabular}{lr}
\toprule
\textbf{Metric} & \textbf{Value} \\
\midrule
GPU VRAM             & 8.2\,GB \\
Precision            & bfloat16 (no quantization) \\
Trainable parameters & 52M / 2{,}152M (2.4\%) \\
Model load time      & 1.3\,s \\
VRAM at load / peak  & 4.27\,GB / 4.61\,GB \\
Avg inference latency & 5.7\,s per frame \\
Avg throughput       & 42\,tok/s \\
English output rate  & 100\% (8/8 question types) \\
\bottomrule
\end{tabular}
\end{center}
\end{table}

\subsection{Design Space Ablation}

A 42-condition ablation on an earlier pipeline version shaped three
architectural choices (full details in~\cite{b20}).
Arrow-style radar overlays improved zero-shot spatial accuracy by +128\%
(2B) and +449\% (8B) but caused $-$15\% degradation after fine-tuning due
to geometric conflict with learned cluster shapes, motivating the
filled-wedge redesign.
Appending ground-truth detection text achieved the highest zero-shot
result (SVA 30.6\%) but caused complete failure when removed at inference,
validating the anti-shortcut schema.
The zero-shot 8B peaks at 15.3\% spatial accuracy; the fine-tuned 2B
reaches 13.8\% on clean LiDAR, a pattern confirmed at larger scale in the
main judge evaluation.

\section{Discussion}
\label{sec:discussion}

Before reading the results, it is worth being precise about what MoRAL
does and does not do.
It is not a perception system: it does not detect objects, estimate depth,
or run a sensor fusion backbone.
What it does is read a pre-rendered spatial image and answer physics
questions about what it sees.
The gains in Table~\ref{tab:perq} are gains in that reading-and-reasoning
task, not in detection accuracy or closed-loop driving.
A fine-tuned 8B on the same data would probably score higher; we did not
test that.
What we can say is that the BEV reading ability, not the parameter count,
is what separates Condition~C from Condition~D.
The EMERGENCY\_BRAKE recall gap says it directly: the 8B model misses
89.2\% of critical cases not because it lacks language capacity, but
because it cannot decode the wedge size and direction that tell it an
object is closing fast.

Condition~C is the only condition with empirically validated BEV reading
before Stage~2 evaluation; Conditions B and D reflect language-model priors
conditioned on a partially unread sensor image.
The primary metric relies on a Gemma judge calibrated on 40--80 frames
rather than a full multi-rater expert study; a second independent
multimodal judge and a larger stratified annotation would strengthen
validity.
The Condition~A vs.\ B/C/D comparison changes both modality type and count,
serving as a performance floor rather than a controlled ablation.
Future work will extend to additional datasets, explore integration with
3D BEV perception pipelines to test generalization of grounded reasoning
across BEV construction methods, and incorporate temporal multi-frame
reasoning, which should substantially improve TTC estimation and far-range
accuracy.

The over-reaction bias observed in Condition~C has a direct physical
deployment implication.
A system that generates false positive EMERGENCY\_BRAKE outputs---predicting
imminent collision when none exists---would trigger unnecessary hard braking
in a real vehicle, risking rear-end collisions and degrading passenger
comfort to the point of driver override.
In Fig.~\ref{fig:confusion}, 37--48\% of non-emergency GT rows in
Condition~C are assigned EMERGENCY\_BRAKE, compared to 9--13\% in
Condition~D.
This phantom braking rate makes Condition~C unsuitable for unsupervised
deployment even in an advisory role, and motivates two directions:
(1)~calibrating the decision threshold using precision--recall tradeoffs on
a held-out safety split rather than maximizing recall alone, and
(2)~conditioning EMERGENCY\_BRAKE predictions on temporal confirmation
across consecutive frames, which should suppress single-frame false
positives without substantially degrading true recall.

\section{Future Work}
\label{sec:future}

\textbf{Evaluation framework.}
The current judge setup works but is not rigorous enough to be a public
benchmark.
The next step is pulling in objective metrics from existing datasets---
nuScenes-QA action accuracy, TTC error, threat ranking precision---so the
results are comparable across systems without relying on a single judge
call.
A stratified expert annotation and a second independent multimodal judge
are the immediate priorities.
\textbf{Self-improving training loop.}
The bigger goal is a training pipeline that finds its own failure cases.
Responses that parse correctly but are physically wrong, actions that are
in the right direction but wrong magnitude, uncertainty outputs that do not
track real sensor gaps---these should feed directly back into the next
training stage.
Adding temporal context (3--5 consecutive frames) is probably the single
highest-leverage change; it would let the model observe acceleration
directly rather than inferring it from a single Doppler snapshot~\cite{b22}.
\textbf{Robots and humanoid platforms.}
Nothing about the BEV pipeline is vehicle-specific.
Any platform with LiDAR and range sensors produces the same kind of data.
The more interesting angle is using MoRAL as a monitoring layer on top of
an existing black-box VLA policy: rather than replacing the policy, the
grounded reasoning chain can flag when the policy's implied spatial
understanding contradicts what the sensors show, and trigger a correction
without retraining the whole system.

\section{Conclusion}
\label{sec:conclusion}

MoRAL demonstrates that staged BEV grounding enables metric spatial
reasoning in a compact 2B VLM that zero-shot models at the tested scales
cannot match.
Stage~1 establishes BEV vocabulary learnability (Zone~F1 = 0.89,
near-range ring accuracy 0.69--0.79) while zero-shot baselines cannot
decode it.
Stage~2 produces MoRAL winning seven of eight question types over a
zero-shot 8B baseline, with the largest gains on multi-step physics
reasoning question types.
Emergency braking recall improves from 10.8\% to 47.8\%, degeneration
falls from 94.1\% to 20.8\%, and the pipeline runs on a consumer 8\,GB
GPU at 42\,tok/s without quantization.
The system is not deployment-ready: most critical cases remain undetected
and closed-loop performance is unvalidated.
What this gives the community is a starting point---a compact model that
can actually read sensor-encoded space, with all training code, checkpoints,
and evaluation artifacts public.\footnote{\url{https://huggingface.co/AmbarishGK/moral-v4-nuscenes}}

\section*{Acknowledgment}

The authors thank Dr.\ Bernardo Flores and Dr.\ Vidyacharan Bhaskar for
their feedback, and the nuScenes team for the publicly available dataset.

\smallskip\noindent\textit{AI tool disclosure:}
The first author used Claude (Anthropic)~\cite{b21} to assist with
condensing and restructuring a conference paper draft that the authors had
already written, drawing on the companion thesis~\cite{b20} to ensure the
distilled version preserved the core technical claims without over- or
under-representing the results.
Claude also assisted with portions of the training and evaluation pipeline
code.
All research design, experimental execution, result analysis, thesis
writing, and scientific conclusions are the authors' own work.
The final manuscript was reviewed and edited by the authors.

\end{document}